\documentclass[10pt,twocolumn,letterpaper]{article}

 \usepackage{wacv}              

\usepackage{graphicx}
\usepackage{amsmath}
\usepackage{amssymb}
\usepackage{booktabs}
\usepackage{multirow}
\usepackage{caption}
\usepackage{multirow}
\usepackage{hhline}
\usepackage{tabularray}
\usepackage[accsupp]{axessibility}

\usepackage[pagebackref,breaklinks,colorlinks]{hyperref}

\usepackage[capitalize]{cleveref}
\crefname{section}{Sec.}{Secs.}
\Crefname{section}{Section}{Sections}
\Crefname{table}{Table}{Tables}
\crefname{table}{Tab.}{Tabs.}

\def\wacvPaperID{469} 
\def\confName{WACV}
\def\confYear{2024}

\begin{document}

\title{You Can Run but not Hide: Improving Gait Recognition with Intrinsic Occlusion Type Awareness}

\author{Ayush Gupta, Rama Chellappa\\
Johns Hopkins University\\
3400 North Charles Street, Baltimore, MD, 21218\\
{\tt\small \{agupt120, rchella4\}@jhu.edu}
}
\maketitle

\begin{abstract}
      While gait recognition has seen many advances in recent years, the occlusion problem has largely been ignored. This problem is especially important for gait recognition from uncontrolled outdoor sequences at range - since any small obstruction can affect the recognition system. Most current methods assume the availability of complete body information while extracting the gait features. When parts of the body are occluded, these methods may hallucinate and output a corrupted gait signature as they try to look for body parts which are not present in the input at all. To address this, we exploit the learned occlusion type while extracting identity features from videos. Thus, in this work, we propose an occlusion aware gait recognition method which can be used to model intrinsic occlusion awareness into potentially any state-of-the-art gait recognition method. Our experiments on the challenging GREW and BRIAR datasets show that networks enhanced with this occlusion awareness perform better at recognition tasks than their counterparts trained on similar occlusions.
\end{abstract}

\vspace{-7mm}
\section{Introduction}
\label{sec:intro}
\vspace{-1mm}

Gait-based recognition and identification techniques focus on analyzing and recognizing an individual's walking pattern, and can be very helpful when other identifiers like face are not available. Every person has a characteristic gait pattern\cite{gait-is-unique} consisting of various bio-mechanic features like stride length, walking speed, body posture, and arm swing\cite{biomechanics}. 
These features can be explicitly measured through wearable sensors\cite{wearable-sensors}; however, this is not practical in remote biometrics applications. Thus, vision-based gait recognition has become more popular\cite{deep-gait-survey, deep-gait-survey2} because of its non-intrusive nature and its potential to identify subjects at a distance. 

There have been many indoor datasets for vision-based gait recognition\cite{casia-b, oumvlp}. However, performance on these has started to saturate\cite{deep-gait-survey2, gaitgl} and focus is shifting towards the outdoor, in-the-wild gait recognition task\cite{grew-dataset, briar-dataset}. Such uncontrolled sequences pose many challenges to gait recognition, such as severe turbulence, changes in illumination, extreme altitudes and viewpoints. Because the camera can be at a large distance from the subject, the quality of the image can not be guaranteed in such in-the-wild data. One of the important challenges in uncontrolled sequences is occlusion, where the input itself may be partially blocked due to an obstruction between the subject and the camera. Gait recognition identifies subjects by observing the correlations between the motion of different body parts, and when some of these body parts are blocked from view, it can be difficult to extract a unique gait signature. We aim to address this problem of occlusion in this work.

\begin{figure}[t]
    \centering
    \includegraphics[width=0.95\linewidth]{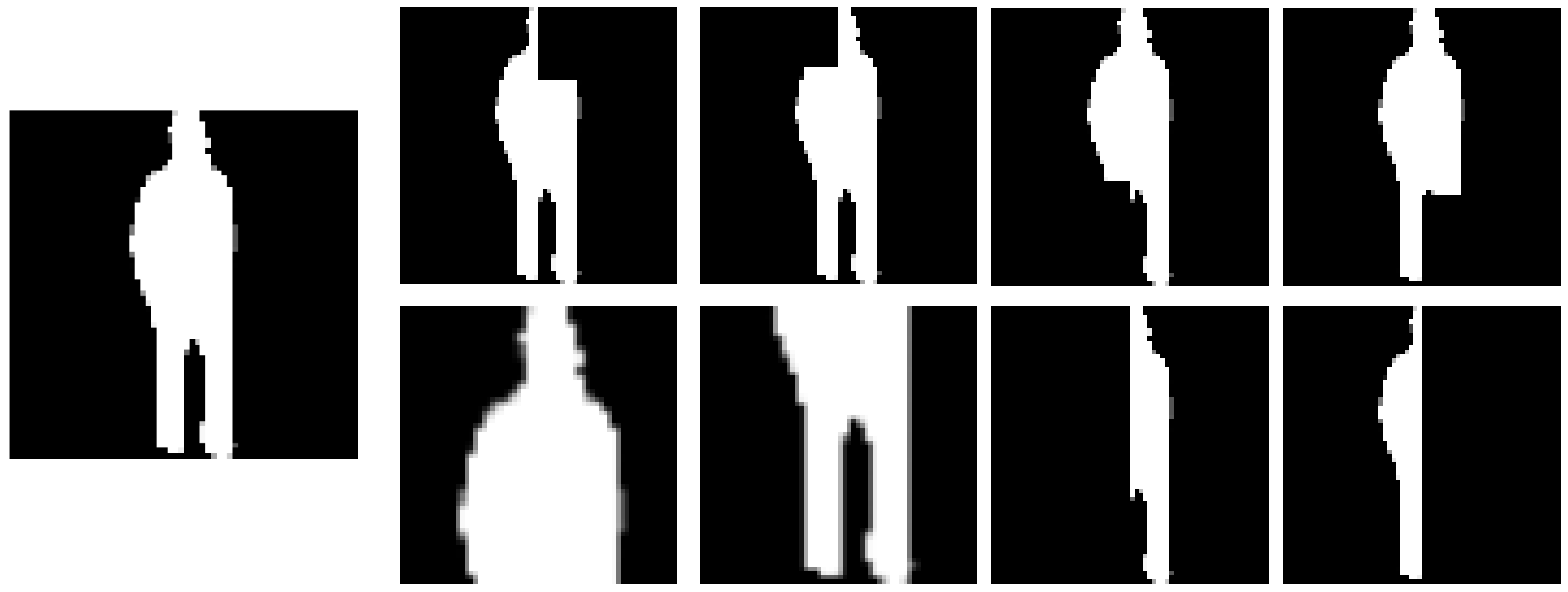}
    \caption{\label{fig:synthetic-occ} Examples of synthetic occlusions used in our method. The leftmost image is an original frame taken from the BRIAR\cite{briar-dataset} dataset, and the other images show the synthetic occlusions applied. We use eight types of synthetic occlusions; placing black patches on any corner of the frame (\#1-\#4), removing the top or bottom part of the frame (\#5-\#6), or blocking out the left or right half of the frame(\#7-\#8). The size and position of the patch is decided randomly for each video. The subject has consented to the use of these images.}
\end{figure}

Occlusion has been relatively well studied in the Person Re-ID field, with large scale datasets specifically targeting occlusion\cite{occluded-reid-dataset, partial-reid-dataset, partial-ilids-dataset}. With only one occluded gait dataset\cite{tumgaid-dataset}, that too captured in controlled conditions, occlusion has not yet gained the attention it deserves within the gait recognition community. Due to the lack of a large scale outdoor occluded dataset, many methods either work on small data collected on their own\cite{gait-occ-dataset} or simulate occlusions on holistic datasets\cite{gait_occ_mesh, spatio_temporal_reid, gait_occ_binary}.  

Current gait recognition techniques assume the availability of the complete body to extract features and extract correlations between the motion of different body parts. When the frame is partially occluded, these correlations get corrupted as the model looks for missing body parts. To address this issue, some attempts have been made for developing algorithms for occluded gait recognition using generative models\cite{gait_occ_binary, spatio_temporal_reid} which complete the occluded frames. However, generative models can often hallucinate unnecessary or irrelevant details. \cite{gait_occ_mesh} uses 3D human mesh modeling and predicts the locations of the occluded body parts by further computations on the SMPL\cite{smpl} features. However, 3D mesh modeling is not reliable from low-quality data captured in uncontrolled environments,  which can further corrupt the predictions.

To tackle these issues, we propose a novel method for occluded gait recognition with intrinsic occlusion type awareness. Our method can be integrated into potentially any state-of-the-art gait recognition model to enhance performance in occluded scenarios. We argue that occlusion type awareness is important to generate discriminative gait signatures from partial, occluded inputs. Without such occlusion awareness, the gait recognition model would try to look for body parts which might not even be present in the input at all - leading to the corruption of the gait signature. Thus, we propose an auxiliary occlusion detection module, which is able to inject occlusion relevant information into the network. The occlusion detector is trained for the occlusion classification task on a variety of different occlusions as shown in Figure \ref{fig:synthetic-occ}. This information is later used to guide the gait recognition backbone.
Our experiments show that embedding such occlusion aware encodings into the network improves the gait recognition performance on challenging outdoor datasets like BRIAR\cite{briar-dataset} and GREW\cite{grew-dataset}.

In summary, our main contributions are as follows:

\begin{itemize}
    \item  We propose a novel model-agnostic, occlusion aware method for gait recognition in uncontrolled environments which can be used with potentially any state-of-the-art method to enhance performance under occlusion.
    \item We design an auxiliary \textit{occlusion detection module} to generate occlusion encodings and show that they contain useful information which can facilitate learning of effective discriminative features for occluded gait recognition.
    \item Through extensive experimentation, we explore various approaches to effectively utilize this occlusion information to optimize model training under occlusion scenarios.
\end{itemize}


\section{Related Work}
\vspace{-1mm}

\textbf{Gait Recognition} has traditionally been done using wearable sensors\cite{wearable-sensors} which capture various aspects of human movement like speed, stride length and body posture. With advances in computer vision techniques, focus has shifted to vision-based methods where gait recognition is performed directly using videos of the subject captured from camera sensors at a distance\cite{deep-gait-survey, deep-gait-survey2, skeleton1, skeleton2, gaitgl, fan_gaitpart_2020}. Within vision based methods, most fall under the category of either skeleton-based\cite{skeleton1, skeleton2} or appearance-based\cite{gaitgl, fan_gaitpart_2020}. Skeleton-based methods extract human keypoints and use the motion of these keypoints to extract features. They rely on the performance of pose estimation techniques, which may not work as well on low-quality data captured from long range. Appearance-based methods, which generally rely on silhouette masks, employ a sequence of 2D and 3D CNNs to capture spatio-temporal information and extract gait-relevant features. Appearance-based methods usually outperform skeleton-based methods\cite{skeleton2}, and they have almost saturated performance on indoor datasets like CASIA-B\cite{casia-b}. More challenging outdoor datasets like GREW\cite{grew-dataset} and BRIAR\cite{briar-dataset} have been collected to further advance research in this domain. \cite{yuxiang-fg23} shows that existing models do not achieve desirable performance when applied directly to these datasets because of challenges  like atmospheric turbulence, changes in illumination and occlusions unique to these datasets.

\textbf{Occluded Person Re-ID} has been a relatively well-studied field, with large scale datasets and techniques\cite{occluded-reid-dataset, simulated_occ_reid, spatio_temporal_reid, aonet, occ_them_all}. Some works simulate different types of occlusions in clever ways to train their networks\cite{simulated_occ_reid, occ_them_all}, while others generate body-part based representations to handle cases when some parts are missing \cite{body-part-based-reid}. However, most of these methods do not utilize any information about the type of occlusion in the input. Some works like \cite{aonet} identify the missing body parts using an occlusion awareness score as part of their architecture, which can work for an image but can not be easily extended for the video modality where the occlusion can change dynamically. On a related note, \cite{quality-aware-net} computes a quality metric for multiple frames and uses this information to combine frame-wise features. However, the model itself does not become quality aware since this information is used only in the aggregation step.

Person Re-ID methods can work well when we have just a single image, but gait recognition  relies on temporal information if it is available. Hence, we discuss Occluded Gait Recognition next.  

\textbf{Occluded gait recognition} has not seen much progress, especially for outdoor, in-the-wild scenarios, mainly due to the lack of a large scale dataset targeting the problem.
Most methods work using synthetic occlusions\cite{spatio_temporal_reid, gait_occ_mesh, gait_occ_binary, occlude-gait-subspace, fdei, ortells2017gait}. Some attempts have been made to complete the occluded portions of the input sequence using generative models\cite{gait_occ_binary, spatio_temporal_reid}. However, generative models can often hallucinate unnecessary and irrelevant details. \cite{occlude-gait-subspace} generates a training matrix of gait features for different occlusions and uses it during verification. \cite{gait_occ_binary} performs binary classification on the input sequence to detect any occlusions and later reconstructs the Gait Energy Image\cite{gei}. However, the type of occlusion is not considered during reconstruction. Works like \cite{fdei, ortells2017gait} try to generate better GEI variants for occlusion, but fail to utilize the temporal information effectively in the process of generating the GEI. Some other methods\cite{gait_occ_mesh} identify and predict the positions of missing body parts in the input using 3D mesh modeling, but they require high quality data which is not always available in outdoor scenes; subsequent prediction of missing keypoints from a noisy mesh model may lead to more errors.

To tackle the shortcomings of existing methods, we present an occlusion aware approach for performing gait recognition, which can be used with potentially any state-of-the-art method to enhance performance in outdoor, in-the-wild conditions. We describe our method in detail in the next section.

\begin{figure*}
    \centering
    \includegraphics[width=0.75\textwidth]{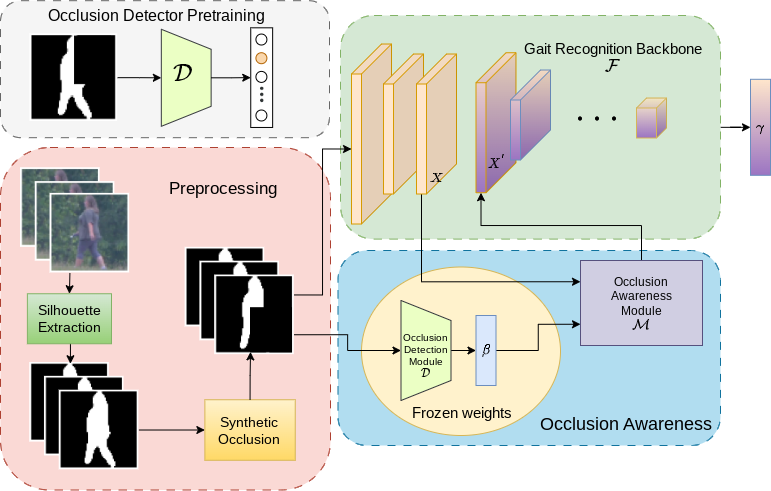}
    \caption{\label{fig:main} An overview of the proposed approach. First, silhouette masks are extracted from input video frames in the preprocessing step. Next, the video is synthetically occluded. This video is passed through the occlusion detection module to give occlusion aware embeddings $\beta$, which are subsequently used to guide the gait recognition backbone $\mathcal{F}$ in generating occlusion aware gait signatures $\gamma$. It is important to note that the occlusion detection module is frozen during training of $\mathcal{F}$ to preserve its occlusion awareness.
    }
\end{figure*}

\section{Method}
\vspace{-2mm}

The main aim of this work is to show that occlusion-guided training yields better results when training a gait recognition network on occluded data. Given a (possibly occluded) video sequence $V^i = \{v_1, v_2, ... v_f\}$  of size $f \times H \times W$ for subject $i$, the goal is to find an identifying gait signature $\gamma_i$ for the subject which may be used for downstream tasks. 

Our approach can be divided into four main steps: 1) The preprocessing stage, where we extract silhouette masks and introduce synthetic occlusions in the videos; 2) The occlusion detector pretraining step, where we train the occlusion detector $\mathcal{D}$ on the classification task with synthetic occlusions; 3) The gait recognition backbone $\mathcal{F}$, which extracts gait features from the input; and 4) The Occlusion Awareness Module $\mathcal{M}$, which generates occlusion aware features and inserts them into the backbone. An overview of these components is described in the next section.

\vspace{-1mm}
\subsection{Overview}
\vspace{-1mm}
An overview of the proposed method is shown in Figure \ref{fig:main}. To ensure that only gait is captured, we first preprocess $V^i$ to generate silhouette masks if the original video belongs to RGB or grayscale modality. Next, we pass the masked video to the occlusion detector $\mathcal{D}$, which outputs the occlusion aware feature $\beta$ containing information about the type of occlusion in the video. In a parallel branch, the gait recognition backbone $\mathcal{F}$ is also fed the preprocessed video $V^i$. The occlusion feature $\beta$ is inserted in the intermediate layers of $\mathcal{F}$ to guide the feature generation and generate an occlusion aware gait signature $\gamma$.   

\subsection{Preprocessing and Synthetic Occlusions}
\label{sec:synthetic-occlusions}
\vspace{-1mm}
To ensure that only gait is captured, we extract silhouette masks from the RGB videos using Detectron2\cite{wu2019detectron2} for the BRIAR dataset which has RGB videos. We adopt the widely used method of centering, cropping and resizing these masks as done in \cite{opengait}.

We train the occlusion detector and the gait recognition backbone on various kinds of synthetic occlusions, as shown in Figure \ref{fig:synthetic-occ}. To simulate occlusion, consistent black patches are placed on the input frames of a video. To better model practical scenarios, the amount of occlusion is chosen randomly between a range $R$. Where the frame's top or bottom half is completely blocked out, we resize the visible part of the subject to the complete frame to simulate how the occluded subject would be detected during real occlusions. Overall, we use eight broad categories of synthetic occlusions; placing a black patch on any of the four corners of the frame (\#1-\#4), removing a random portion of the top or bottom part of the frame (\#5-\#6), or removing the left or right half of the frame (\#7-\#8). These eight categories of occlusion along with the non-occluded category give a total of nine classes the occlusion detector $\mathcal{D}$ is trained on. While it is possible to train the occlusion detector on more occlusions classes, we feel that it is not practical to train it on every possible occlusion type. As described in Section \ref{sec:occlusion-detection-module}, we discard the classification head of the occlusion detector and use only the penultimate features which can contain relevant information even for unseen occlusions. 

\subsection{Occlusion Detection Module} 
\label{sec:occlusion-detection-module}
\vspace{-1mm}
The occlusion detector $\mathcal{D}$ is a convolutional neural network which is trained to classify the type of occlusion in the input video $V^i$. It consists of three convolutional layers and two additional linear layers, with the final layer classifying the video into nine classes - for eight types of occlusions and the non-occluded case. The network is trained on this classification task using the cross entropy loss. 

The occlusion detector is trained separately from the gait recognition backbone. During its training, the classifier head is used for loss calculation. However, when $\mathcal{D}$ is used to guide the training of the gait recognition backbone $\mathcal{F}$ in the next step, this classifier head is removed. It is important to note that the weights of the occlusion detector are frozen when $\mathcal{F}$ is being trained, to ensure that $\mathcal{D}$ does not lose its intrinsic occlusion awareness.

When guiding $\mathcal{F}$, the occlusion detector can operate in two modes - 1) Cumulative occlusion detection and 2) Transient occlusion detection, depending on the position of inserting of $\beta$ in the backbone. In the cumulative mode, the occlusion detector outputs a single occlusion feature for the whole video. In the transient mode, the occlusion detector outputs different occlusion features for each frame in the input $V$. In summary, the occlusion detection module performs the following computation:
\vspace {-1mm}
\begin{equation}
\label{eqn:occ-detector}
    \beta = \mathcal{D}(V^i)
\end{equation}
where $\beta$ is the occlusion aware feature containing information about the occlusion type in the video $V^i$. If $\mathcal{D}$ is operating in the transient mode, the features are denoted by $\beta_t$, which consists of the output of $\mathcal{D}$ for every frame. On the contrary, if $\mathcal{D}$ is operating in the cumulative mode, the features are represented by $\beta_c$ and denote the output of the occlusion detector averaged across the temporal dimension. In our experiments, we observe that the position where we insert $\beta_c$ into the backbone utilizes occlusion awareness more effectively than the position we insert $\beta_t$.  

\subsection{Gait Recognition Backbone}  
\vspace{-1mm}
The Gait Recognition backbone $\mathcal{F}$ can potentially be any state-of-the-art network for gait recognition. Though state-of-the-art models can take occluded videos as input without any change, they assume the visibility of the entire body. As a result, the performance is impacted when complete gait information is not present since the network is not able to capture the correlations of movement between the visible and occluded body parts. 

To address this problem, we use the occlusion detector $\mathcal{D}$ to guide the training process of $\mathcal{F}$.  Occlusion aware features are inserted into the forward pass of $\mathcal{F}$ so that it can identify the missing body parts and calculate the correlations between the visible and occluded portions accordingly. The occlusion aware features are provided by the occlusion awareness module $\mathcal{M}$, as described in the next section.

\subsection{Occlusion Awareness Module}
\label{sec:occ-awareness-module}
\vspace{-1mm}

The occlusion awareness module $\mathcal{M}$ is used to insert occlusion information into the gait recognition backbone $\mathcal{F}$. It takes the output of the occlusion detector $\beta$, and combines it with intermediate features in $\mathcal{F}$; this is used to guide the gait recognition backbone. Both $\mathcal{F}$ and $\mathcal{M}$ are trained together end to end, while weights of the occlusion detector $\mathcal{D}$ are frozen.
The generation of occlusion aware features is described by the following equation: 
\begin{equation}
\label{eqn: occ-awareness-module}
    X^{'} = \mathcal{M}(X, \beta)
\end{equation}

where $X$ is the intermediate feature computed by $\mathcal{F}$, $\beta$ is the occlusion awareness feature from $\mathcal{D}$, and $X^{'}$ is the new occlusion aware intermediate feature which is replaced by $X$ in the backbone. 
Thus, occlusion information is injected into the forward pass of $\mathcal{F}$ by the occlusion awareness module $\mathcal{M}$ according to Equations \ref{eqn:occ-detector} and \ref{eqn: occ-awareness-module}.

We experiment with different functions for $\mathcal{M}$ as described in Section \ref{sec:ablations}, namely adding $\beta$ to $X$, concatenating $X$ and $\beta$ and passing them through a learnable transformation, and choosing different positions of $X$ from the backbone. We find that introducing a learnable layer in the intermediate layer of $\mathcal{F}$ to insert $X^{'}$, which we call the Deferred Concat method, yields the best performance. 

\begin{figure}[h]
    \centering

    \begin{subfigure}[b]{0.9\linewidth}
        \centering
        \includegraphics[width=\textwidth]{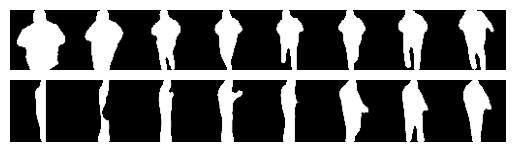}
    \end{subfigure}

    \caption{\label{fig:sample-images} Some sample frames taken from the GREW\cite{grew-dataset} dataset. Only the silhouette masks have been publicly released due to privacy concerns.}
\end{figure}

\begin{figure}[h]
    \centering
    \includegraphics[width=0.9\linewidth]{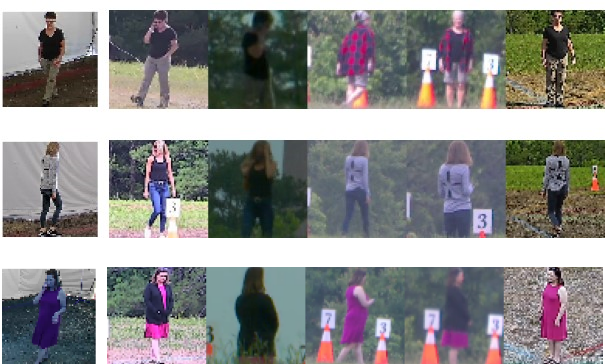}
    \caption{Examples of original video frames taken from the BRIAR\cite{briar-dataset} dataset. The leftmost column shows images taken from the indoor sequences used in the gallery set. Next columns have images taken from 100m, 200m, 400m, 500m, and elevated angle respectively for the same subject. Illumination changes and turbulence make the task challenging.
    These subjects have consented to use of these images in publication.}
    \label{fig:briar-rgb-samples}
\end{figure}

\section{Experiments}

\subsection{Datasets}
\label{sec:data}
\vspace{-1mm}

\textbf{BRIAR}\cite{briar-dataset} is a recently collected outdoor dataset containing many challenging variations in clothing, illumination conditions, viewpoints, distances and altitudes. Videos are captured at multiple distances from different camera sensors, including some aerial videos. Some videos are captured in controlled environments from high quality cameras as well, and these are used as gallery sequences during evaluation. The outdoor videos captured from long range and UAVs are of low quality because of atmospheric turbulence and are used as the probe set. Furthermore, the clothing of the subjects is different in the gallery and probe. Our subset of the BRIAR dataset contains a total of 212 training subjects and 90 test subjects, both sets being mutually exclusive. Some examples of the original frames taken from BRIAR are presented in Figure \ref{fig:briar-rgb-samples}. Additional details have been included in the supplementary material.

\textbf{GREW}\cite{grew-dataset} is a large scale publicly available dataset for gait recognition, comprising of 20,000 subjects in the training split and 6,000 in the testing split. The dataset is captured from in-the-wild videos in a variety of conditions from multiple different cameras with varying viewpoints. Only the preprocessed silhouettes are publicly available due to privacy concerns. Some example frames from the GREW dataset are shown in Figure \ref{fig:sample-images}.

\subsection{Data preprocessing}
\vspace{-1mm}
To ensure that only gait is used and appearance features are filtered out, we calculate segmentation masks using Detectron 2\cite{wu2019detectron2} on all the input frames for the BRIAR dataset. We crop, centre and resize all these silhouette masks to a standard $H \times W = 64 \times 64$  size. 

For the GREW dataset, we follow the widely adopted preprocessing technique as in \cite{opengait}, where we crop the images to include only the bounding box around the subject and resize it to 64 $\times$ 64. Since the GREW dataset is already in the form of silhouette masks, we do not run any segmentation model on it. 

\subsection{Backbones and Baselines}
\vspace{-1mm}
We experiment with three different state-of-the-art models for our backbone $\mathcal{F}$, namely GaitGL\cite{gaitgl}, GaitPart\cite{fan_gaitpart_2020} and GaitBase\cite{opengait}. All of these are CNN-based networks. We use the implementations of these models from the OpenGait\cite{opengait} repository. All backbones are trained using a part-wise triplet loss as done in \cite{opengait}, with GaitGL and GaitBase having an additional cross-entropy loss.

We perform experiments with two different types of baselines as mentioned in Tables \ref{tab:main-briar} and \ref{tab:main-grew}. Baseline-1 is when we train the backbone $\mathcal{F}$ on complete videos of the subjects without any occlusion, and perform zero-shot evaluation on occluded sequences. Baseline-2 has these same architectures which are trained and evaluated on similar kind of synthetically occluded data.

\subsection{Implementation Details}

\vspace{-1mm}
The gait recognition backbone $\mathcal{F}$ accepts a video of silhouette masks. We set the size of the input frames to $64 \times 64$. During training, we also fix the number of input frames $f$ to 30 during training. The occlusion detection module outputs an occlusion aware embedding $\beta$ for the video. Depending on the mode of operation of the occlusion detection module, $\beta$ can be a single 1-D vector of size 64 (in the cumulative mode), or a 2-D vector of size $f \times 64$, where $f$ is the number of frames in the input (in the transient mode). $\beta$ is later fed into $\mathcal{F}$ at different positions as described in Section \ref{sec:ablations}.

We use the PyTorch\cite{pytorch} framework to conduct our experiments. We use the backbone models provided in the OpenGait\cite{opengait} repository. We train our baselines and occlusion aware models with the Adam optimizer\cite{adam-optim}, a learning rate of 0.0001 and a batch size of $(8, 8)$ for training GaitGL and GaitPart and $(32, 4)$ for GaitBase.

\subsection{Evaluation}

\textbf{BRIAR}\cite{briar-dataset} dataset provides the evaluation protocol in terms of the gallery and probe set split. The gallery set consists of high quality controlled videos captured indoors, while the probe set contains more challenging outdoor conditions. More details about the evaluation protocol have been included in the supplementary material. We compute the top-$K$ rank retrieval accuracy at different distances for evaluation on BRIAR.

For the \textbf{GREW}\cite{grew-dataset} dataset, we follow the standard protocol mentioned in Section \ref{sec:data}, and follow \cite{opengait} for computing top-$K$ rank retrieval accuracy. 

\textbf{Multiple evaluations: }
Since our evaluation results are based on randomly chosen occlusions, we repeat the evaluation ten times and report the mean and the standard deviation across these multiple runs to check the effect of randomness in our main experiments.

\section{Results and Discussion}
\subsection{Occlusion Detector}
The occlusion detector is trained separately from the backbone $\mathcal{F}$ for classifying the occlusion type. It is able to identify the correct occlusion type with 98\% accuracy on the BRIAR test set. We only train it on the relatively smaller BRIAR dataset and use it directly for our experiments with GREW to demonstrate its robustness across domains for the task of occlusion detection. Additional cross-domain  evaluation results for the occlusion detector are included in the supplementary material.

\begin{table*}
\centering
\resizebox{\linewidth}{2cm}{%
\begin {tabular}{c||c|cc|cc|cc|cc|cc}
\multirow{2}{*}{Backbone} & \multirow{2}{*}{Method} & \multicolumn{2}{c|}{100m}                         & \multicolumn{2}{c|}{400m}                         & \multicolumn{2}{c|}{500m}                         & \multicolumn{2}{c|}{Elevated Angle}                & \multicolumn{2}{c}{Aerial}                         \\ 
\cline{3-12}
                          &                         & Rank-1                  & Rank-20                 & Rank-1                  & Rank-20                 & Rank-1                  & Rank-20                 & Rank-1                  & Rank-20                 & Rank-1                  & Rank-20                  \\ 
\hline
\multirow{3}{*}{GaitBase\cite{opengait}} & Baseline-1              & 16.34 $\pm$ 1.934          & 50.34 $\pm$ 2.538          & 13.72 $\pm$ 1.263          & 48.98 $\pm$ 2.940          & 10.96 $\pm$ 1.422          & 46.33 $\pm$ 3.582          & 14.06 $\pm$ 1.394          & 48.17 $\pm$ 3.329          & 14.76 $\pm$ 7.680          & 49.84 $\pm$ 15.409          \\
                          & Baseline-2              & 24.50 $\pm$ 1.738          & 83.59 $\pm$ 1.661          & 20.00 $\pm$ 2.070          & 78.67 $\pm$ 1.829          & 16.62 $\pm$ 1.925          & 75.39 $\pm$ 2.128          & 24.60 $\pm$ 1.277          & 79.30 $\pm$ 1.242          & 23.81 $\pm$ 6.809          & 71.59 $\pm$ 7.185           \\ 
\cline{2-12}
                          & Occlusion Aware         & \textbf{31.99 $\pm$ 2.593} & \textbf{86.63 $\pm$ 1.591} & \textbf{24.79 $\pm$ 1.875} & \textbf{80.89 $\pm$ 1.665} & \textbf{23.38 $\pm$ 1.765} & \textbf{78.60 $\pm$ 1.915} & \textbf{29.58 $\pm$ 0.865} & \textbf{82.32 $\pm$ 0.837} & \textbf{32.38 $\pm$ 5.226} & \textbf{78.57 $\pm$ 5.418}  \\ 
\hhline{=::===========}
\multirow{3}{*}{GaitGL\cite{gaitgl}}   & Baseline-1              & 12.76 $\pm$ 1.434          & 63.24 $\pm$ 2.094          & 6.43 $\pm$ 0.984           & 53.39 $\pm$ 3.338          & 6.62 $\pm$ 1.439           & 51.74 $\pm$ 2.130          & 9.12 $\pm$ 0.843           & 56.94 $\pm$ 1.091          & 9.37 $\pm$ 2.879           & 59.84 $\pm$ 7.206           \\
                          & Baseline-2              & 11.58 $\pm$ 1.186          & 65.60 $\pm$ 2.586          & 5.19 $\pm$ 1.357           & 50.21 $\pm$ 2.281          & 4.31 $\pm$ 0.828           & 50.17 $\pm$ 2.219          & 7.52 $\pm$ 0.640           & 59.11 $\pm$ 2.024          & 4.60 $\pm$ 3.363           & 65.08 $\pm$ 10.836          \\ 
\cline{2-12}
                          & Occlusion Aware         & \textbf{34.66 $\pm$ 1.549} & \textbf{81.81 $\pm$ 1.485} & \textbf{20.14 $\pm$ 1.366} & \textbf{70.00 $\pm$ 1.729} & \textbf{16.90 $\pm$ 1.546} & \textbf{70.91 $\pm$ 2.204} & \textbf{26.72 $\pm$ 1.506} & \textbf{75.29 $\pm$ 1.232} & \textbf{26.83 $\pm$ 8.413} & \textbf{81.27 $\pm$ 3.237}  \\ 
\hhline{=::===========}
\multirow{3}{*}{GaitPart\cite{fan_gaitpart_2020}} & Baseline-1              & 4.65 $\pm$ 0.717           & 45.41 $\pm$ 1.186          & 2.32 $\pm$ 0.414           & 34.49 $\pm$ 1.472          & 2.21 $\pm$ 0.673           & 35.03 $\pm$ 1.519          & 3.85 $\pm$ 0.661           & 43.01 $\pm$ 1.201          & 4.60 $\pm$ 3.131           & 50.63 $\pm$ 5.958           \\
                          & Baseline-2              & 12.55 $\pm$ 1.424          & 73.12 $\pm$ 1.871          & 5.59 $\pm$ 0.940           & 51.18 $\pm$ 2.515          & 5.14 $\pm$ 1.230           & 48.81 $\pm$ 1.970          & 9.14 $\pm$ 0.617           & 64.49 $\pm$ 2.397          & 10.16 $\pm$ 5.079          & 77.14 $\pm$ 5.776           \\ 
\cline{2-12}
                          & Occlusion Aware         & \textbf{21.17 $\pm$ 1.913} & \textbf{78.80 $\pm$ 1.956} & \textbf{11.58 $\pm$ 1.146} & \textbf{66.91 $\pm$ 1.420} & \textbf{11.57 $\pm$ 1.303} & \textbf{65.35 $\pm$ 2.302} & \textbf{15.59 $\pm$ 0.977} & \textbf{72.72 $\pm$ 1.400} & \textbf{11.59 $\pm$ 5.769} & \textbf{75.87 $\pm$ 6.098} 
\end{tabular}
}
\caption{\label{tab:main-briar}
Comparison of the baselines and occlusion aware networks on the BRIAR\cite{briar-dataset} dataset on synthetic occlusions. Rank-$K$ retrieval accuracies averaged across ten evaluation runs are reported for different ranges along with the standard deviation. Baseline-1 is zero shot evaluation on occluded data, Baseline-2 is the network which has seen synthetic occlusions during training. We can see the training on synthetic occlusions with occlusion awareness performs the best for occluded data. 
}
\end{table*}

\begin{table*}
\centering
\resizebox{0.7\linewidth}{!}{%
\begin{tabular}{c||c|cccc}
Backbone                  & Method                   & Rank-1                  & Rank-5                  & Rank-10                 & Rank-20                  \\ 
\hline
\multirow{3}{*}{GaitBase\cite{opengait}} & Baseline-1               & 15.15 $\pm$ 0.263          & 25.34 $\pm$ 0.437          & 30.67 $\pm$ 0.465          & 36.41 $\pm$ 0.385           \\
                          & Baseline-2               & 27.26 $\pm$ 0.554          & 42.66 $\pm$ 0.478          & 49.81 $\pm$ 0.425          & 57.08 $\pm$ 0.476           \\ 
\cline{2-6}
                          & \textbf{Occlusion Aware} & \textbf{34.57 $\pm$ 0.522} & \textbf{50.73 $\pm$ 0.486} & \textbf{57.93 $\pm$ 0.488} & \textbf{64.82 $\pm$ 0.396}  \\ 
\hhline{=::=====}
\multirow{3}{*}{GaitGL\cite{gaitgl}}   & Baseline-1               & 5.95 $\pm$ 0.327           & 10.40 $\pm$ 0.313          & 12.97 $\pm$ 0.316          & 15.78 $\pm$ 0.370           \\
                          & Baseline-2               & 10.59 $\pm$ 0.177          & 20.73 $\pm$ 0.319          & 26.96 $\pm$ 0.435          & 34.40 $\pm$ 0.523           \\ 
\cline{2-6}
                          & \textbf{Occlusion Aware} & \textbf{13.77 $\pm$ 0.280} & \textbf{25.91 $\pm$ 0.383} & \textbf{32.70 $\pm$ 0.506} & \textbf{40.60 $\pm$ 0.594}  \\ 
\hhline{=::=====}
\multirow{3}{*}{GaitPart\cite{fan_gaitpart_2020}} & Baseline-1               & 5.18 $\pm$ 0.237           & 9.99 $\pm$ 0.375           & 12.76 $\pm$ 0.359          & 16.00 $\pm$ 0.396           \\
                          & Baseline-2               & 8.71 $\pm$ 0.323           & 18.11 $\pm$ 0.423          & 23.79 $\pm$ 0.442          & 30.56 $\pm$ 0.399           \\ 
\cline{2-6}
                          & \textbf{Occlusion Aware} & \textbf{13.77 $\pm$ 0.441} & \textbf{25.84 $\pm$ 0.329} & \textbf{32.42 $\pm$ 0.484} & \textbf{39.90 $\pm$ 0.627} 
\end{tabular}
}
\caption{\label{tab:main-grew}Comparison of the baselines and the occlusion aware networks on the GREW\cite{grew-dataset} dataset after introducing synthetic occlusions. Rank retrieval accuracies averaged across ten evaluation runs are reported along with the standard deviation. We can see that occlusion awareness helps across different backbones.}
\end{table*}

\subsection{Occlusion Aware Gait Recognition}

Our results are summarized in Tables \ref{tab:main-briar} and \ref{tab:main-grew}. Baseline-1, which refers to zero-shot evaluation on occluded sequences, generally performs the worst since the model has never seen occluded sequences during training. Baseline-2 performs slightly better but is still not optimal, since it is difficult for a single network to work on different occlusion types on its own. Our results show that using an auxiliary occlusion detector to introduce occlusion awareness helps the model to generate better identity features under occlusions.

Our results also show that even though occlusions are chosen randomly during evaluation, the results are mostly consistent across multiple evaluations, with the exception of the aerial condition on BRIAR which deviates significantly from the mean. This is because only thirty subjects have aerial videos out of the ninety test subjects, and the small dataset size causes randomness  to become a larger factor in evaluation. 

We also observe that Baseline-1 performs better than Baseline-2 for the GaitGL backbone. Since these are data-driven methods, we do not have a proper explanation for this behavior; this may be because it is difficult to optimize the slightly larger GaitGL model on the turbulent BRIAR data with synthetic occlusions. However, the occlusion aware network still performs better than both the two baselines.

\begin{table*}
\centering
\resizebox{0.8\linewidth}{!}{
\begin{tabular}{c||cc|cc|cc|cc|cc}
\multirow{2}{*}{Occlusion
  Awareness Method} & \multicolumn{2}{c|}{100m} & \multicolumn{2}{c|}{400m} & \multicolumn{2}{c|}{500m} & \multicolumn{2}{c|}{Extreme Angle} & \multicolumn{2}{c}{Aerial}  \\
                                              & Rank1 & Rank20            & Rank1 & Rank20            & Rank1 & Rank20            & Rank1 & Rank20                     & Rank1 & Rank20              \\ 
\hline
Guided Add~                                   & 21.31 & 82.76             & 12.36 & 70.47             & 13.38 & 71.97             & 16.65 & 73.55                      & 15.87 & 60.32               \\
Learnable 3DConv                              & 27.3  & 81.05             & 14.15 & 72.39             & 13.38 & 74.31             & 21.37 & 73.55                      & 19.05 & 82.54               \\
3D Conv + Deferred Concat                     & 15.52 & 73.23             & 7.28  & 58.79             & 5.1   & 52.65             & 9.8   & 65.54                      & 3.17  & 74.6                \\
Complex Deferred Concat                       & 32.44 & 79.87             & 16.48 & 67.03             & 14.86 & 66.45             & 25.91 & 75.6                       & 31.75 & 87.3                \\ 
\hline
Deferred Concat                               & 34.58 & 82.12             & 21.15 & 70.19             & 18.47 & 70.91             & 25.73 & 78.27                      & 28.57 & 82.54              
\end{tabular}
}
\caption{\label{tab:briar-analysis}A comparison of different methods of injecting occlusion awareness into the gait recognition backbone, on the BRIAR\cite{briar-dataset} dataset. We observe that the Deferred Concat method performs best and adding complexity to the occlusion awareness model may help in some cases. Also, we see that too much occlusion information can distract the network as observed from the 3D Conv + Deferred Concat row.}
\end{table*}

\begin{table}
\centering
\resizebox{\linewidth}{!}{
\begin{tabular}{c||cccc}
Occlusion
  Awareness Method & Rank-1         & Rank-5         & Rank-10        & Rank-20         \\ 
\hline
Guided Add~                  & 7.73           & 16.27          & 21.57          & 28.23           \\
Learnable 3DConv             & 8.13           & 17.47          & 22.92          & 29.77           \\
3D Conv + Deferred Concat        & 11.73          & 22.05          & 28.78          & 35.88           \\
Deferred Concat                  & 13.9           & 26.2           & 32.68          & 40.68 \\
\hline
Complex Deferred Concat          & 14.25 & 26.48 & 33.08 & 40.3            
\end{tabular}
}
\caption{\label{tab:grew-analysis}A comparison of different occlusion awareness methods on the synthetically occluded GREW dataset. Too much occlusion information distracts the network as seen from the third row. The Complex Deferred Concat method works better here because of the larger size of the GREW dataset.}
\end{table}

\subsection{Analysis and Ablations}
 \label{sec:ablations}

 We perform various experiments to analyze the performance of the model with different occlusion awareness methods and on different occlusions. All experiments in this section are done using the GaitGL backbone on the GREW dataset, unless stated otherwise.

\vspace{-3mm}
 
 \subsubsection{Optimal location for occlusion awareness}
 \vspace{-2mm}
 In this section, we experiment with variants of the occlusion awareness module $\mathcal{M}$ to identify the optimal method for introducing occlusion awareness in the backbone and producing the features $X^{'}$. The results are presented in Tables \ref{tab:briar-analysis} and \ref{tab:grew-analysis}, and are discussed in this section.

\textbf{Guided Add: }The first variant we tried adds the outputs of the occlusion detector directly to the second convolution layer of the CNN-based backbone, as described by: 
\vspace{-1mm}
\begin{equation}
    X^{'} = X + \beta_{t}
\end{equation}
\vspace{-1mm}
where $\beta_t$ means that the occlusion detector operates in the transient mode, generating a $\beta$ feature for each frame. However, we observe that this method does not perform well. Even though $\beta_t$ contains occlusion information, it may not be directly compatible to be used with the backbone features $X$ as they come from different networks. Hence, we introduce a learnable transformation over these features in the next experiment to make the features compatible.

\textbf{Learnable 3DConv: }In this variant,
instead of adding $X$ and $\beta_t$, they are concatenated across the channel dimension and passed through a learnable 3D convolutional layer.

It is important to note that the occlusion detector is still not trained, but an additional transformation layer is introduced so that both $X$ and $\beta_t$ become compatible, resulting in better performance. This method is described by:

\vspace{-3mm}

\begin{equation}
    X^{'} = \mathcal{M}( X \oplus \beta_t )
\end{equation}
where $\oplus$ denotes the concatenation operation across the channel dimension and the occlusion awareness module $\mathcal{M}$ is a learnable 3D convolutional layer.

\textbf{Deferred Concat: }Here, we experiment with a different position to introduce occlusion awareness in the backbone. Specifically, we introduce the occlusion aware features at the \textit{Head0} layer of GaitGL and GaitPart, and FCs layer of GaitBase, as defined in the OpenGait repository\cite{opengait}. These layers are one of the deeper layers of the network, which is why we call this method Deferred Concat. We follow a similar procedure as Learnable 3D Conv, concatenating the output across the channel dimension as shown by: 

\vspace{-3mm}
\begin{equation}
    X^{'} = \mathcal{M}(X \oplus \beta_c)
\end{equation}
\vspace{-2mm}

where $X$ are the features taken from the deeper \textit{Head0} or \textit{FCs} layer and $X^{'}$ are the obtained occlusion aware features. Since these are flat layers, $\mathcal{M}$ is also a learnable fully connected layer. Since this layer does not contain any temporal information, the occlusion detector operates in the Cumulative mode in this method, generating a single occlusion feature for the entire video denoted by $\beta_c$. We observe that this performs better than Learnable 3D Conv, and we think it is because occlusion awareness may be more useful in the later layers when the network looks at the global scale of the image, instead of the earlier low level layers.

\textbf{Deferred Concat + 3D Conv: }In this experiment, we inject occlusion information at multiple points in the backbone $\mathcal{F}$. To do this, we combine both the above two methods, namely Learnable 3D Conv and Deferred Concat. We observe that performance actually deteriorates compared to using just the Deferred Concat method, as the model can get distracted with too much occlusion information.

\textbf{Complex Deferred Concat:  }In this experiment, we add an additional linear layer to the deferred concat awareness module $\mathcal{M}$ to increase the complexity of the occlusion aware features $X^{'}$. For the GREW dataset, we observe that this improves performance, but not for the BRIAR dataset. We think this is because the larger GREW dataset requires more complex representations of occlusion awareness because of its larger size, but the complex model may overfit on the relatively smaller BRIAR dataset.

\vspace{-3mm}
\subsubsection{Optimal location across backbones}
\vspace{-2mm}
All experiments in the previous section were performed using the GaitGL\cite{gaitgl} backbone. Here, we experiment with another backbone to see if the observations about the optimal location for occlusion awareness hold true across different gait recognition models. Thus, we perform another experiment with GaitPart\cite{fan_gaitpart_2020}, the results of which are shown in Table \ref{tab:gaitpart-occ-position}. We observe that the Deferred Concat method still works better. We conjecture that for arbitrary CNN backbones, inserting occlusion aware features would work better in the later layers since the model can extract local level features in the earlier layers and consolidate these with occlusion awareness in the later layers.

\begin{table}
\centering
\resizebox{0.8\linewidth}{!}{%
\begin{tabular}{c|cccc}
                  & Rank-1 & Rank-5 & Rank-10 & Rank-20  \\ 
\hline
Learnable 3D Conv & 12.45  & 25.07  & 31.60    & 39.40     \\
Deferred Concat   & 13.93  & 26.00     & 32.73   & 40.33   
\end{tabular}
}
\caption{\label{tab:gaitpart-occ-position} Inserting occlusion awareness in different positions in the GaitPart backbone, evaluated on GREW. We observe that even in GaitPart, occlusion awareness is more useful in the deeper layers using the Deferred Concat method. }
\end{table}

\vspace{-3mm}
\subsubsection{Generalization to unseen occlusions}
\vspace{-2mm}
We train and evaluate our models on a fixed set of broad occlusion types, since we feel that training a model on all possible occlusion types is not practical. We try to verify whether this approach generalizes to different occlusions by evaluating these trained models directly on unseen occlusion types. Specifically, we evaluate our approach on synthetic dynamic occlusions and also on real jagged occlusions present in the BRIAR dataset. Descriptions about these new occlusion types have been included in the supplementary material. Our results, presented in Table \ref{tab:unseen-occ}, show that our approach can generalize to unseen occlusions as well. This is because even though $\mathcal{D}$ is trained on nine occlusion classes, its penultimate layer can still contain occlusion relevant information about which body parts are present in the input. This occlusion awareness can be utilized by the backbone to generate better features.

\begin{table}
\centering
\resizebox{\linewidth}{0.7cm}{%
\begin{tabular}{c|cccc||cc}
\multirow{2}{*}{Method} & \multicolumn{4}{c||}{Synthetic Dynamic Occlusions} & \multicolumn{2}{c}{Real Jagged
  Occlusions}  \\ 
\cline{2-7}
                        & Rank-1 & Rank-5 & Rank-10 & Rank-20               & Rank-1 & Rank-20                                      \\ 
\hline
Baseline - 2              & 20.75  & 34.00  & 39.87   & 45.85                 & 4.94   & 51.33                                        \\
Occlusion aware         & 25.18  & 40.2   & 46.37   & 52.32                 & 23.86  & 85.78                                       
\end{tabular}
}
\caption{\label{tab:unseen-occ} Generalizability to unseen occlusions with the GaitGL backbone with and without occlusion awareness. Synthetic dynamic occlusions are applied on GREW and real jagged occlusions are used from BRIAR.}
\end{table}

\section{Limitation and Future Work}
\vspace{-2mm}

We demonstrate that occlusion type awareness helps in generating more discriminative features for occluded gait recognition. Our current method identifies the occlusion type and passes along the information to guide the gait recognition backbone. We believe that introducing changes within the backbone architecture itself will help further since their current design  assumes the availability of the full body. Additionally, we mostly deal with synthetic occlusions in this work, and it is still not clear how well these synthetic occlusions can simulate real world occlusions. As such, datasets specifically addressing the occlusion problem in gait recognition are needed to further advance research in this area. Lastly, we only experiment with gait recognition methods on silhouettes. We believe that exploiting all modalities and combining Person Re-ID and face recognition methods with gait recognition can improve recognition performance and we leave this to future work.

\vspace{-2mm}
\section{Conclusion}
\vspace{-2mm}
Existing gait recognition methods are not effective when occlusions are present in the input. Thus, in this work, we propose a model-agnostic approach to introduce \textit{intrinsic occlusion awareness} into existing models to enhance their performance in occlusion scenarios. We perform experiments on the BRIAR and GREW datasets, both of which contain uncontrolled outdoor sequences. We train a domain-robust auxiliary occlusion detector and use it to inject occlusion aware features into three different state-of-the-art gait recognition backbones. We find that occlusion awareness is best used in the intermediate layer of the backbones. We compare our method with two different baselines and conclude that intrinsic occlusion awareness can help in increasing performance on occluded gait recognition.

\section{Acknowledgement}
\vspace{-2mm}
This research is based upon work supported in part by the Office of the Director of National Intelligence (ODNI), Intelligence Advanced Research Projects Activity (IARPA), via [2022-21102100005]. The views and conclusions contained herein are those of the authors and should not be interpreted as necessarily representing the official policies, either expressed or implied, of ODNI, IARPA, or the U.S. Government. The US. Government is authorized to reproduce and distribute reprints for governmental purposes notwithstanding any copyright annotation therein.

{\small
\bibliographystyle{ieee_fullname}
\bibliography{egbib}
}

\maketitlesupplementary

\section{Introduction}
\label{sec:supple-intro}

In this supplementary material, we provide more details about the BRIAR dataset and its evaluation protocol. Next, we elaborate on the synthetic occlusions used in our experiments. We further train new models on dynamic occlusions and show that occlusion awareness can help even in dynamic occlusions. Next, we provide additional details and analysis on the Learnable 3D Conv method. Further, we perform experiments with different occlusion types, and restrict occlusion types in the gallery and probe set to analyse how difficult different occlusions types are. Lastly, we provide more details and experimental evaluation results regarding the occlusion detector.

\section{BRIAR Dataset}

The BRIAR\cite{briar-dataset} dataset is a recently collected dataset for gait recognition in outdoor, uncontrolled conditions. It has a lot of challenging outdoor scenes containing large variations in illumination, camera quality, distance of the subject from the camera, and extreme viewpoints. This makes it one of the most challenging gait recognition datasets. 

The dataset contains videos captured systematically from distances of 100m, 200m, 400m and 500m. Additionally, some videos are captured from UAVs and some are captured at close range from an elevated viewpoint. Some video frames captured from UAVs are visualized in Figure  \ref{fig:uav-vis}. In BRIAR, the subjects move inside a fixed square boundary. The movement of the subjects may be 1) structured, where they walk along pre-defined straight lines inside the boundary, or 2) random, where subjects can move arbitrarily inside the boundary. While walking, the subjects are free to use their mobile phones and walk naturally, to represent a more practical scenario. 

We use the BRS-1 and BRS-1.1 subsets of the BRIAR dataset for training, giving us a total of 212 training subjects. We use the BTS-1 subset for evaluation, containing 90 subjects. The BRS and BTS subsets are mutually exclusive, so no subjects used for training are used for evaluation and vice versa. The dataset defines the protocol for evaluation, containing the subject IDs, and start and end frame for each of the probe and gallery sequences.

Additionally, the videos captured from the 200m range are kept at a position which introduces jagged occlusions where the lower part of the subject is always occluded from view from tall grass of varying height. This makes recognizing the gait especially difficult, since the legs are partially hidden from view of the camera and the occlusion is also not consistent across the video. Some examples of these jagged occlusions have been shown in Figure 4 of the main paper.

\subsection{Evaluation Protocol}

The BRIAR dataset contains a variety of different conditions and distances. We take the non-occluded videos from the dataset and introduce synthetic occlusions in them for evaluation. The BRIAR protocol provides the probe and gallery split. A single video may have multiple probes within it, as specified by the start and end frames of each probe according to the protocol. It should be noted that none of the probes overlap with each other. The BRIAR dataset also contains single images as probes, but we filter them out because temporal information is required to run gait recognition models. 

The controlled, indoor sequences in BRIAR are of higher quality and are treated as the gallery set. Meanwhile, the outdoor, more challenging conditions constitute the probe set. For evaluation, we use the Top-$K$ rank retrieval metric. We compute the euclidean distance between each probe-gallery pair, and select the top $K$ gallery matches for each probe. If the correct identity of the probe subject is within the top $K$ predictions, the subject is regarded as being identified correctly. Since each subject also has multiple entries within the gallery, we select the top $K$ gallery videos instead of the top $K$ subjects. This list may have a subject being repeated, effectively reducing the number of possible candidates to choose from. Thus, this is a tougher evaluation metric than selecting the top $K$ unique subjects while also being a more practical one to evaluate the model on.

\section{Synthetic Occlusions}

\subsection{Consistent Occlusions }
We use synthetic consistent occlusions to train the occlusion detector $\mathcal{D}$ as well as the gait recognition backbone $\mathcal{F}$ in most of our experiments. Consistent occlusions are one where all the frames have the occlusion patch at the same position, thereby blocking a body part from view for the entire length of the video. The consistent occlusion types we use in our experiments are described in Figure 1 and Section 3.2 of the main paper. The range $R$ of these synthetic occlusions is set to be 20\% - 50\% of the frame size. The level of occlusion in a video is randomly chosen from this range.

\subsection{Dynamic Occlusions }
Dynamic occlusions are one where the position of the occlusion patch changes with time. We perform some additional experiments using such dynamic occlusions to check the generalizability of the occlusion aware model to unseen occlusion types, and to verify whether using the occlusion detector in the transient mode through the Learnable 3D Conv technique helps with dynamic occlusions.

\begin{figure}
    \centering
    \includegraphics[width = 0.8\linewidth]{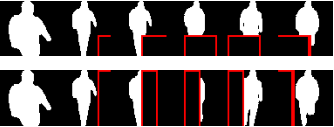}
    \caption{\label{fig:dynamic-occ-vis} Examples of the synthetic dynamic occlusions we use, applied on video frames taken from the GREW dataset. The top row shows a small moving rectangular patch, and the bottom row shows a tall patch which covers the height of the frame applied on the same video. The occlusion patches are shown with a red boundary for visualization purposes only.}
\end{figure}

\begin{figure}
    \centering
    \includegraphics{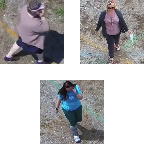}
    \caption{\label{fig:uav-vis}Some sample frames taken from the videos captured from UAVs in the BRIAR\cite{briar-dataset} dataset. We can see the extreme viewpoint angle in these videos, making recognition a more challenging problem from these.} 
\end{figure}

To simulate dynamic occlusions, we place black patches of different shapes on the image frames, and the position of these patches can change with time. Specifically, we place either a small rectangular moving patch which occludes a portion of the subject, or a tall rectangular moving patch which covers the entire height of the frame. Some examples of these dynamic occlusions are shown in Figure \ref{fig:dynamic-occ-vis}. The height and width of the small patch are chosen randomly within the range $R_{ds} = (0.3, 0.5)$ which corresponds to 30\% - 50\% of the frame dimension. The height of the tall rectangular patch of occlusion is fixed to the height of the frame, and its width is chosen randomly within the range $R_{dt} = (0.2, 0.4)$ for each video. We decide these ranges of the occlusion patch size by manually visualizing the occluded video for different ranges, and choosing one which looks most similar to occlusion patterns which might be caused by objects like trees or poles covering the height of the frame, or small stationary objects like cones or boxes blocking a part of the moving subject.

To make the occlusions dynamic and realistic, we decide to give a velocity to the occlusion patches as opposed to randomly deciding the position across each frame. The direction of movement of the patch is decided randomly from left to right or right to left, and the velocity of these patches is chosen randomly from the range $R_v = (0.5, 1.0)$ pixels per frame. This range has also been chosen by manually inspecting the synthetically occluded videos with different velocities for the occlusion patch.

\section{Occlusion Awareness in dynamic occlusions}

Here, we train occlusion aware networks on dynamic occlusions. We experiment with Learnable 3D Conv and the Deferred Concat method to insert occlusion awareness in the GaitGL backbone. The results are summarized in Table \ref{tab:dynamic-occ-result}. We observe that even in dynamic occlusions, the Deferred Concat method, where $\mathcal{D}$ operates in cumulative mode and outputs $\beta_c$, performs better than Learnable 3D Conv where $\mathcal{D}$ operates in transient mode and outputs $\beta_t$. This further demonstrates that the position where Learnable 3D Conv inserts occlusion aware features is not optimal for the gait recognition backbone, and occlusion awareness helps in the deeper layers of the network. It remains to be seen whether inserting transient occlusion features $\beta_t$ into these deeper layers further improves performance on dynamic occlusions, and we leave that to future work.

\begin{table}
\centering
\resizebox{0.9\linewidth}{!}{%
\begin{tabular}{c|cccc}
Method            & Rank-1 & Rank-5 & Rank-10 & Rank-20  \\ 
\hline
Learnable 3D Conv & 22.45  & 37.45  & 44.42   & 51.93    \\
Deferred Concat   & 30.32  & 47.08  & 54.45   & 62.62   
\end{tabular}
}
\caption{\label{tab:dynamic-occ-result} Different occlusion awareness methods used for dynamic occlusions. Even for occlusions which change with time, inserting occlusion information in the deeper layers of the network performs better.
}
\end{table}

\section{Learnable 3D Conv}
\subsection{Additional details}
The occlusion awareness module $\mathcal{M}$ takes as input the occlusion feature $\beta$ and the intermediate feature $X$. It outputs a new occlusion aware intermediate feature $X^{'}$ which is replaced by $X$ in the backbone. In most of the experiments, the size of $X^{'}$ is same as $X$, so that the architecture of the backbone remains unchanged. However, in Section \ref{sec:larger-size}, we experiment with a larger size of the intermediate feature $X^{'}$. 

The size of the occlusion feature $\beta$ is $64 \times f$ in transient mode ($f$ being the number of frames in the video), and $64 \times 1$ in cumulative mode. In the Learnable 3D Conv method, the transient occlusion feature $\beta_t$ is repeated along height and width dimensions and concatenated with the intermediate feature $X$ (of size $32 \times f \times h \times w$) along the channel dimension to give a feature size of $96 \times f \times h \times w$. The learnable 3D Conv layer reduces this again to $32 \times f \times h \times w$. However, the 3D Conv described in Section \ref{sec:larger-size} transforms it into another block of $96 \times f \times h \times w$.     

\subsection{Increasing number of channels in 3D Conv}

\label{sec:larger-size}
In this experiment, we try a larger size of the intermediate feature $X^{'}$ to see if a larger size of the occlusion feature benefits the model. Specifically, we use the Learnable 3D Conv method and increase the number of output channels in the 3D Conv to 96 from the earlier 32 channels. We use the GaitGL\cite{gaitgl} backbone and train the model on the BRIAR\cite{briar-dataset} dataset. The results are mentioned in Table \ref{tab:bigdim}. We compare the model to the earlier Learnable 3D Conv and the Deferred Concat method, and observe that increasing the number of channels actually hurts the model. Thus, the occlusion information is able to fit better in the original number of channels of $X^{'}$ and introducing more channel confuses the model.

\begin{table*}
\centering
\resizebox{\linewidth}{!}{
\begin{tabular}{c||cc|cc|cc|cc|cc}
\multirow{2}{*}{Occlusion
  Awareness Method} & \multicolumn{2}{c|}{100m} & \multicolumn{2}{c|}{400m} & \multicolumn{2}{c|}{500m} & \multicolumn{2}{c|}{Extreme Angle} & \multicolumn{2}{c}{Aerial}  \\
                                              & Rank1          & Rank20   & Rank1          & Rank20   & Rank1          & Rank20   & Rank1          & Rank20            & Rank1          & Rank20     \\ 
\hline
Learnable 3D Conv more channels                    & 22.27          & 77.94    & 9.48           & 62.36    & 9.77           & 57.96    & 14.25          & 66.79             & 12.7           & 71.43      \\
Learnable 3D Conv                              & 27.3           & 81.05    & 14.15          & 72.39    & 13.38          & 74.31    & 21.37          & 73.55             & 19.05          & 82.54      \\ 
\hline
Deferred Concat                               & \textbf{34.58} & 82.12    & \textbf{21.15} & 70.19    & \textbf{18.47} & 70.91    & \textbf{25.73} & 78.27             & \textbf{28.57} & 82.54     
\end{tabular}
}
\caption{\label{tab:bigdim} Comparison of the performance of the occlusion aware network when the intermediate feature has more channels, compared to the regular Learnable 3D Conv method. The best deferred concat method is also shown for reference. We observe that increasing the number of channels actually hurts the performance, and the occlusion information better fits in the original number of channels inside the backbone.}
\end{table*}

\section{Occlusion Type Analysis}

\paragraph{Evaluation by occlusion type: }
During evaluation, we randomly apply occlusions of different types on the input. In this section, we evaluate our model on these occlusion types separately to get an idea about which occlusion types are easier and which are difficult for the model. Our results are summarized in Table \ref{tab:eval-by-occ-type}. As expected, the model is able to perform better when the size of the occlusion patch is small (corresponding to occlusions \#1-\#4). However, the task becomes much more difficult when half of the body is missing in occlusions \#5-\#8.

\begin{table}
\centering
\resizebox{\linewidth}{!}{%
\begin{tabular}{c|cccc}
Occlusion
  types         & Rank-1 & Rank-5 & Rank-10 & Rank-20  \\ 
\hline
Corner patch (\#1-\#4)    & 20.92  & 35.12  & 42.13   & 48.82    \\
Half Horizontal (\#5-\#6) & 7.97   & 14.35  & 17.87   & 21.8     \\
Half vertical (\#7-\#8)   & 11.67  & 21.37  & 27.9    & 36.35   
\end{tabular}
}\caption{\label{tab:eval-by-occ-type} Evaluation of the occlusion aware GaitGL model on the GREW dataset, where synthetic occlusions are restricted to particular types during evaluation. \#1-\#4 correspond to an occlusion patch placed in any of the four corners of the frame, \#5-\#6 correspond to half horizontal occlusions where the top or bottom half of the body may be missing, and \#7-\#8 correspond to occlusions where either the left or the right half of the body may be missing. Half horizontal is the toughest occlusion type and corner patches are relatively the easiet occlusion types for the model.}
\end{table}

\paragraph{Different occlusions in gallery and probe: }
In our experiments, the occlusion type for each video is chosen at random, independent of other videos. As a consequence, the gallery and probe videos of a subject may have different or the same type of occlusion within them. In this section, we analyse the effect of enforcing the gallery and probe set to have different occlusions. As such, we apply occlusion types \#1-\#4 on the gallery set, and restrict occlusions on the probe set to \#5-\#6 and \#7-\#8 in separate experiments. 
Our results are shown in Table \ref{tab:gall-probe-diff-occ}. Here as well, we observe that the model with occlusion awareness is able to perform better than the baselines. 

\begin{table*}
\centering
\resizebox{0.9\linewidth}{!}{%
\begin{tabular}{c|cc|cc|cc|cc}
\multirow{2}{*}{Method} & \multicolumn{2}{c|}{Rank-1} & \multicolumn{2}{c|}{Rank-5} & \multicolumn{2}{c|}{Rank-10} & \multicolumn{2}{c}{Rank-20}  \\ 
\cline{2-9}
                        & P \#5-\#6 & P~ \#7-\#8      & P \#5-\#6 & P~ \#7-\#8      & P \#5-\#6 & P~ \#7-\#8       & P \#5-\#6 & P~ \#7-\#8       \\ 
\hline
Baseline-1              & 0.23      & 1.27            & 0.6       & 3.23            & 1.18      & 4.8              & 2.33      & 7.03             \\
Baseline-2              & 9.8       & 15.3            & 19.67     & 28.95           & 25.2      & 35.78            & 31.5      & 43.85            \\
Occlusion Aware         & 11.4      & 18.22           & 21.68     & 33.62           & 27.88     & 40.7             & 35.05     & 47.8            
\end{tabular}
}
\caption{\label{tab:gall-probe-diff-occ}Performance of the baselines and occlusion aware model when gallery and probes have different occlusion types. Here, gallery occlusion is chosen between \#1-\#4 and probe occlusion is chosen from either \#5-\#6 or \#7-\#8 as specified.}
\end{table*}

\section{Occlusion Detector}

The occlusion detector $\mathcal{D}$ takes a video of silhouette masks as input and outputs the occlusion feature $\beta$. It is trained on silhouette images to classify the image into nine classes - eight types of occlusions or no occlusion. When working on videos, it outputs an occlusion feature for every frame and depending on its mode of operation, it can either output the entire block $\beta_t$ or the mean-pooled feature $\beta_c$.

\paragraph{Optimal architecture: } In our experiments, we use a three-layer convolutional neural network as the occlusion detector. In this section, we try out different depths of the CNN architecture to see which one would be the best for introducing occlusion awareness. We try out networks with 1, 3 and 5 convolutional layers, and they are able to achieve classification accuracies of 89.1\%, 98.8\% and 99.1\% respectively. Even though the 5 layer network performs the best, the difference in performance is not much between the latter two variants. Thus, we choose the 3 layer network to introduce occlusion awareness considering the trade-off between computational cost and performance.

\paragraph{Implementation Details: }
The occlusion detector $\mathcal{D}$ we use is a three layer CNN with two additional linear layers. The ReLU activation function is used after each layer, except the last one where we use the softmax activation function while training. The occlusion detector is trained on the occlusion classification task using Cross Entropy Loss\cite{cross-entropy} and the Adam optimizer\cite{adam-optim} with a learning rate of 0.001. We use a batch size of 32 for training the occlusion detector. The occlusion detector is trained on the BRIAR dataset, from which silhouette masks are extracted using Detectron2\cite{wu2019detectron2}.

\subsection{Training and evaluation}
During training, we sample one frame from every 50 frames in the video. During evaluation of the occlusion detector, we randomly pick one frame from each video. During both training and evaluation, synthetic occlusions of the previously discussed eight types are randomly introduced during the data loading step of the input frame and the classification accuracy is measured.

\subsection{Architecture}
The architecture of the occlusion detector is described in Table \ref{tab:occ-detector-arch}. It is a three-layer convolutional neural network followed by two linear heads. During training, the output of the FC2 layer is used to calculate the cross-entropy loss. However, during inference, and when it is being used along with the backbone $\mathcal{F}$, the FC2 layer is removed and the output of FC1 is used as the occlusion feature $\beta$.

\begin{table}[ht]
\centering
\resizebox{0.8\linewidth}{!}{%
\begin{tabular}{c|c|c}
\textbf{Layer Name} & \textbf{Input shape} & \textbf{Output Shape}  \\ \hline
Conv1               & 64 * 64 * 1          & 64 * 64 * 32           \\
ReLU, MaxPool1      & 64 * 64 * 32         & 32 * 32 * 32           \\
Conv2               & 32 * 32 * 32         & 32 * 32 * 64           \\
ReLU, MaxPool2      & 32 * 32 * 64         & 16 * 16 * 64           \\
Conv3               & 16 * 16 * 64         & 16 * 16 * 128          \\
ReLU, MaxPool3      & 16 * 16 * 128        & 8 * 8 * 128            \\
AdaptiveAvgPool     & 8 * 8 * 128          & 128                    \\ \hline
FC1                 & 128                  & 64                     \\
FC2                 & 64                   & 9                     
\end{tabular}
}

\caption{\label{tab:occ-detector-arch} The architecture of the occlusion detector. It is a three layer convolutional neural network followed by two fully connected layers.}
\end{table}


\section{Cross Domain Evaluation of Occlusion Detector}


For our experiments with the gait recognition backbone $\mathcal{F}$, we use the weights of the occlusion detector obtained after training it on the relatively smaller BRIAR dataset. We use it directly on GREW without additional training to demonstrate its robustness across different domains. In this section, we further demonstrate its cross-domain generalization capability. We train and evaluate it on both BRIAR and GREW datasets, and also perform cross domain evaluation on the occlusion classifying task.

The results obtained are presented in Table \ref{tab:cross-domain-eval}. We observe that while the model performs best in-domain, the performance does not drop significantly across domains, thus demonstrating the robustness of the occlusion detector $\mathcal{D}$. 

\begin{table}
\centering
\resizebox{0.8\linewidth}{!}{
\begin{tabular}{c|c|c}
               & Test on BRIAR & Test on GREW   \\ 
\hline
Train on BRIAR & \textbf{98.0} & 94.9           \\ 
\hline
Train on GREW  & 97.9          & \textbf{98.8} 
\end{tabular}
}
\caption{\label{tab:cross-domain-eval}In-domain and cross-domain evaluation of the occlusion detector on the BRIAR and GREW datasets. As expected, the performance is highest in the in-domain evaluation, but it does not drop significantly across domains. This demonstrates the robustness of the occlusion detector across domains.}
\end{table}

\end{document}